\documentclass{article} 
\usepackage{arxiv,times}

\usepackage{amsmath,amsfonts,bm}

\def\eqref#1{equation~\ref{#1}}

\def\1{\bm{1}}

\DeclareMathAlphabet{\mathsfit}{\encodingdefault}{\sfdefault}{m}{sl}
\SetMathAlphabet{\mathsfit}{bold}{\encodingdefault}{\sfdefault}{bx}{n}

\usepackage{hyperref}
\usepackage{url}

\usepackage{amsmath,amssymb}
\usepackage{graphicx}
\usepackage{booktabs}
\usepackage{array}
\usepackage{xcolor}
\usepackage{microtype}
\usepackage{caption}
\usepackage{tikz}
\usepackage{subcaption}
\usepackage{wrapfig}
\usepackage{enumitem}

\title{When Routing Reveals Membership: Privacy Leakage from MoE Router Telemetry}

\author{
Yixin Tan$^{1}$ \quad
Jiayang Liu$^{2}$ \quad
Lu Sun$^{3}$ \quad
Yuke Hu$^{4}$ \quad
Zheng Li$^{5}$ \quad
Rui Wen$^{1}$
\\[0.5em]
$^{1}$Institute of Science Tokyo
\\
$^{2}$Nanyang Technological University
\\
$^{3}$Tohoku University
\\
$^{4}$King Abdullah University of Science and Technology
\\
$^{5}$Shandong University
}

\iclrfinalcopy 

\begin{document}

\maketitle


\begin{abstract}
Mixture-of-Experts (MoE) language models produce routing information during inference that may be logged or exposed for monitoring, debugging, load analysis, and safety auditing. 
Unlike ordinary model outputs, this telemetry reveals a view of the model's internal computation, raising a privacy question: can it reveal whether an example was used to fine-tune the deployed model?
We introduce a router-augmented membership inference attack that combines conventional output-side signals with aggregated routing features and applies a membership classifier learned from independently fine-tuned shadow models to the target model.
Across three MoE architectures and three data domains, router telemetry consistently improves membership inference over a strong output-signal ensemble, increasing TPR at 1\% FPR by 2.7--9.4 percentage points across all nine settings. 
The leakage persists across full fine-tuning, frozen-router training, LoRA, and instruction tuning, and remains observable with only discrete expert selections, restricted telemetry, or a single shadow model.
Mechanistic analysis further shows that the leakage does not require router-specific memorization: fine-tuning introduces membership information into hidden representations, while the router exposes a projection of this signal even when its parameters are frozen. 
Perturbing the telemetry reduces this additional leakage only as its fidelity degrades. 
Our results show that router telemetry can turn an operational signal into an additional privacy surface for fine-tuned MoE models.
\end{abstract}

\section{Introduction}
Mixture-of-Experts (MoE) language models route each token through a small subset of expert networks selected by a learned router, producing routing decisions throughout inference~\citep{MOE,OLMoE}. Pretrained MoE models are commonly adapted to downstream tasks through fine-tuning on task- or domain-specific data and subsequently deployed for inference. During deployment, the routing process also produces information about how tokens are assigned across experts, which may be retained or surfaced as telemetry for monitoring, debugging, load analysis, or safety auditing~\citep{MoE_Tutel,lv2026routescannonintrusiveapproachauditing}. Such telemetry exposes signals derived from intermediate representations that are not available from model outputs alone. This raises a natural privacy question: can exposed router telemetry enable membership inference and reveal whether a candidate input was used to fine-tune the deployed model? In this work, we investigate the membership-inference risk associated with router telemetry in deployed MoE language models.

Existing membership inference attacks on language models primarily rely on signals available from model outputs, including likelihoods, confidence statistics, and changes relative to a reference checkpoint~\citep{MIA_Min-K,MIA_Min-K++,WLBZ24,MIA_WBC,MIA_EZ}.
However, membership information may be only weakly reflected in final outputs, limiting the effectiveness of such attacks, particularly at stringent low-false-positive operating points~\citep{das2025blind,duan2024membership}.
Recent studies have therefore explored membership signals encoded in intermediate representations, including hidden states, attention patterns, and gradient-induced feature drift~\citep{MIA_Neural_breadcrumbs,MIA_Attenmia,MIA_G_drift}.
Exploiting these signals, however, typically requires direct access to internal model states, which is unavailable in standard deployment settings.
Router telemetry occupies an intermediate regime: it is derived from internal representations, yet may be exposed as an operational signal during inference.
It may therefore reveal membership information beyond that available from model outputs, without requiring direct access to hidden states.
Whether this creates a meaningful membership-inference risk remains largely unexplored.

\begin{figure*}[t]
    \centering
    \includegraphics[width=0.85\linewidth]{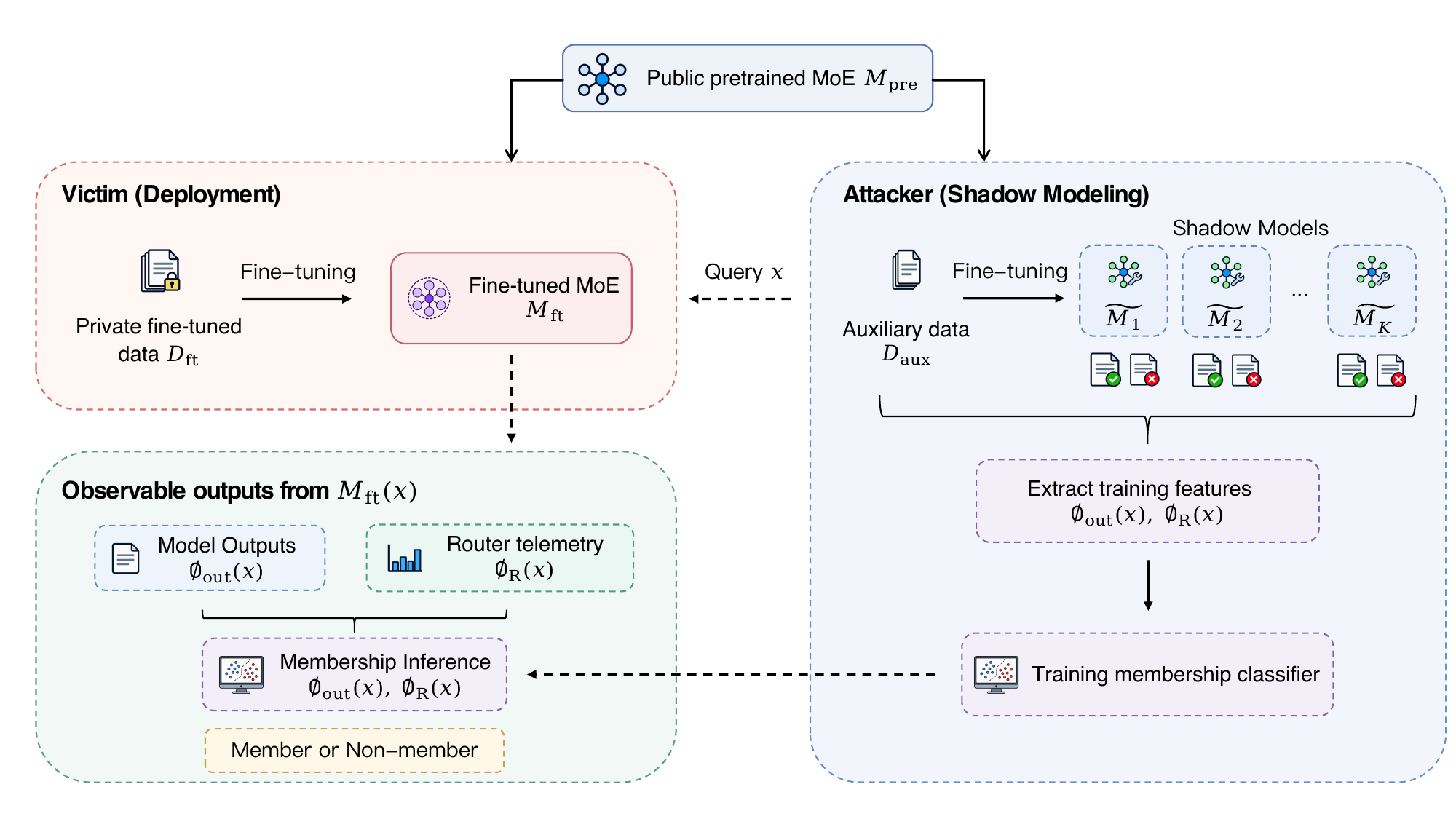}
    \caption{
    Overview of the threat model and router-augmented membership inference pipeline.
    }
    \label{fig:threat_model_overview}
\end{figure*}

Figure~\ref{fig:threat_model_overview} illustrates the deployment setting and router-augmented membership inference pipeline considered in this work.
We consider an attacker who can query a fine-tuned MoE model, observe its outputs and corresponding router telemetry, and access the public pretrained checkpoint as a reference, but has no access to the target model's parameters, hidden states, or private fine-tuning data.
For each candidate input, the attacker combines output-side membership signals with a fixed-dimensional router representation obtained by aggregating the exposed routing information across tokens, experts, and layers.
Since membership labels for the victim's fine-tuning data are unavailable, the attacker independently fine-tunes shadow MoE models on auxiliary data from the same domain and uses their training and held-out examples as labeled members and non-members.
A membership classifier is then trained on the combined output-side and router representations and applied directly to the target model to produce a membership score.

Across three MoE architectures and three data domains, our experiments show that exposed router telemetry consistently increases membership leakage beyond output-side signals, with particularly clear gains at the low-FPR operating point.
The effect persists across different fine-tuning regimes, including full fine-tuning, frozen-router training, LoRA, and instruction tuning.
We further examine how the leakage changes under restricted telemetry exposure and limited shadow-model access, and find that the membership signal remains observable even with partial routing information or only a single shadow model.
Our mechanistic analysis shows that the leakage is not specific to router memorization: membership information is already encoded in fine-tuned hidden representations, while the router exposes a projection of this signal even when its parameters are frozen.
Finally, we evaluate telemetry perturbations as potential defenses and find that reducing the additional leakage comes with a corresponding degradation in telemetry fidelity.

\textbf{Contributions.}
Our contributions are as follows.

\begin{itemize}
    \item \textbf{Router-telemetry membership leakage in deployed MoE models.}
    We identify and systematically study membership leakage through router telemetry exposed during MoE inference, revealing an additional privacy surface beyond conventional model outputs.

    \item \textbf{Router-augmented membership inference.}
    We develop an attack that combines output-side membership signals with aggregated router telemetry and learns the membership classifier from independently fine-tuned shadow models on same-domain auxiliary data.

    \item \textbf{Comprehensive empirical characterization.}
    Across three MoE architectures and three data domains, router telemetry consistently adds membership signal beyond output-side attacks.
    We further show that the leakage persists across fine-tuning regimes and restricted attacker access, and reflects membership information encoded in hidden representations rather than router-specific memorization.
\end{itemize}

\section{Related Work}

\textbf{Membership inference in language models.}
Membership inference attacks (MIAs) aim to determine whether an example was used to train or fine-tune a model~\citep{MIA}.
Most existing attacks derive membership signals from model outputs.
Reference-free methods use likelihood- or confidence-based statistics, including Min-K\%~\citep{MIA_Min-K}, Min-K\%++~\citep{MIA_Min-K++}, ReCaLL~\citep{MIA_ReCaLL}, prediction confidence~\citep{MIA_Confidence}, and in-context optimization signals such as ICP~\citep{MIA_ICP_MIA}.
Reference-based methods instead calibrate target-model outputs using auxiliary models or checkpoints.
LiRA~\citep{MIA_LiRA} estimates member and non-member score distributions with shadow models, \cite{MIA_qantifying} compare candidate likelihoods under target and reference models, and SPV~\citep{MIA_SPV_MIA} builds a reference model from self-prompted data.
For fine-tuned language models, the pretrained checkpoint provides a natural reference.
WBC~\citep{MIA_WBC} compares localized token-loss patterns between the fine-tuned and pretrained models, EZ~\citep{MIA_EZ} measures probability shifts at prediction-error positions, and LTsig~\citep{MIA_LTsig} learns membership signatures from token-level distributional differences between the two models.
Despite these advances, output-based methods are limited to information reflected in the model's predictions and may not capture all membership-related signals encoded during fine-tuning.

A complementary line of work studies membership signals in internal representations.
\cite{MIA_Neural_breadcrumbs} exploit hidden-state and attention features for membership inference, while \cite{MIA_Attenmia} use attention patterns to distinguish members from non-members.
\cite{MIA_G_drift} further exploit gradient-induced changes in internal representations and model outputs to infer membership.
These studies show that internal model states can contain membership information beyond that captured by output statistics.
However, exploiting such signals requires access to hidden states, attention patterns, or other internal model information, which is typically unavailable in deployment settings.

\textbf{Privacy risks of MoE routing.}
Existing work has shown that routing behavior in MoE models can reveal sensitive information about inference-time inputs. 
\cite{yona2024stealinguserpromptsmixture} exploit tie-handling behavior in Expert-Choice Routing to recover victim prompts, while \cite{nuriyev2026expert} show that expert selection traces can reveal substantial information about input tokens. 
\cite{avinash2026task} further demonstrates that routing signatures can predict the task category of an input, and \cite{ding2025moecho} exploit hardware side channels induced by expert execution to infer information about prompts and responses. 
These studies establish that routing-related signals can expose properties of the input being processed, but do not investigate whether router telemetry can reveal information about the model's fine-tuning data. 
More relevant to our setting, GateBleed~\citep{kalyanapu2025gatebleed} demonstrates membership inference through a hardware timing side channel and, separately, shows that MoE expert choices can be inferred from the same hardware signal, which requires hardware-level side-channel access that is typically unavailable to external attackers. 
In contrast, we study the membership-inference risk introduced when router telemetry is exposed in deployed MoE systems.

\section{Problem Statement}
\label{sec:prelim}

\textbf{Mixture-of-Experts models.}
A Mixture-of-Experts (MoE) language model replaces some or all dense feed-forward blocks with a set of $E$ expert networks and a learned router. 
For token $t$ at MoE layer $\ell$, let $h_{\ell,t}\in\mathbb{R}^{d}$ denote its hidden representation. 
The router computes
\[
z_{\ell,t}=W_{\ell}h_{\ell,t}\in\mathbb{R}^{E},
\qquad
p_{\ell,t}=\operatorname{softmax}(z_{\ell,t}),
\]
where $W_{\ell}$ is the router weight matrix and $p_{\ell,t}$ gives the routing distribution over experts. 
The router selects the top-$k$ experts,
\[
S_{\ell,t}=\operatorname{TopK}(p_{\ell,t},k),
\]
and routes the token to the corresponding expert networks.

For an input sequence $x=(x_1,\ldots,x_T)$, we use $R(x)$ to denote the routing information produced during inference. 
Depending on the exposure granularity, $R(x)$ may contain continuous routing scores or probabilities, discrete selected-expert identities, or aggregated expert-usage statistics derived from them. 
We collectively refer to such information as \emph{router telemetry}.

\textbf{Membership inference attacks.}
Let $M_{\mathrm{pre}}$ denote a pretrained model, and let $M_{\mathrm{ft}}$ denote the model obtained by fine-tuning $M_{\mathrm{pre}}$ on a private dataset $D_{\mathrm{ft}}$.
Given a candidate example $x$, membership inference aims to determine whether $x$ was included in $D_{\mathrm{ft}}$. 
We define the membership label as
\[
m(x)=
\begin{cases}
1, & x\in D_{\mathrm{ft}},\\
0, & x\notin D_{\mathrm{ft}}.
\end{cases}
\]

A membership inference attack produces a score $s(x)\in\mathbb{R}$, where a larger value indicates stronger evidence that $x$ is a member.
Given a decision threshold $\tau$, the attack predicts
\[
\hat{m}(x)=\mathbb{I}\!\left[s(x)\geq\tau\right].
\]

\textbf{Metrics.}
For a decision threshold $\tau$, the true-positive rate (TPR) and false-positive rate (FPR) are defined as
\[
\operatorname{TPR}(\tau)
=
\Pr\!\left[s(x)\geq\tau \mid m(x)=1\right],
\]
and
\[
\operatorname{FPR}(\tau)
=
\Pr\!\left[s(x)\geq\tau \mid m(x)=0\right].
\]

We report the area under the receiver operating characteristic curve (\emph{AUROC}) as a threshold-independent measure of membership separability. 
Because membership inference is particularly relevant at low false-positive rates, we additionally report
\[
\operatorname{TPR}@\alpha\%\operatorname{FPR},
\]
defined as the TPR achieved at an operating threshold corresponding to an FPR of $\alpha\%$. 
We use $\operatorname{TPR}@1\%\operatorname{FPR}$ as our primary low-FPR metric.

\subsection{Threat Model}
\label{sec:threat}

\textbf{Victim's behavior.}
The victim fine-tunes a pretrained MoE model $M_{\mathrm{pre}}$ on a private dataset $D_{\mathrm{ft}}$ and deploys the resulting model $M_{\mathrm{ft}}$ as an inference service.
During inference, the serving infrastructure may retain or expose per-request router telemetry $R(\cdot)$ for operational purposes such as monitoring, debugging, or expert-load analysis.

\textbf{Attacker's knowledge.}
The attacker knows the publicly available pretrained checkpoint $M_{\mathrm{pre}}$ from which the target model is fine-tuned, as well as the domain of the downstream fine-tuning data. 
However, the attacker does not know the victim's private fine-tuning dataset $D_{\mathrm{ft}}$. The attacker has no access to the parameters, intermediate hidden states, or gradients of $M_{\mathrm{ft}}$.

\textbf{Attacker's capability.}
The attacker observes token-level output probabilities (or logits) returned by the deployed model together with the corresponding router telemetry.
The attacker can also query the public checkpoint $M_{\mathrm{pre}}$ to construct reference-based attack signals.
In addition, the attacker can independently fine-tune one or more copies of $M_{\mathrm{pre}}$ on auxiliary data from the same domain and use these shadow models to train a membership classifier that is applied to the target model.

\textbf{Attacker's goal.}
Given a candidate example $x$, the attacker's goal is to determine
whether
\[
x \in D_{\mathrm{ft}},
\]
i.e., whether $x$ was used to fine-tune the deployed target model.

\section{Router-Augmented Membership Inference}
\label{sec:method}

Our attack augments conventional output-side membership inference with
router telemetry exposed by the target MoE model. For each candidate
example, we combine output-derived membership signals with a
fixed-dimensional representation of its routing behavior. The
membership classifier is learned from independently fine-tuned shadow
models constructed using auxiliary data from the same domain, and is
then applied to the target model.

\paragraph{Output-side membership signals.}
We construct an ensemble of existing output-side membership signals to
characterize the membership information available from conventional
model outputs. The ensemble combines both target-model signals derived
from $M_{\mathrm{ft}}$ and reference-based signals obtained by comparing
$M_{\mathrm{ft}}$ with the public pretrained checkpoint
$M_{\mathrm{pre}}$. For an input $x$, we denote the resulting
output-side representation as
\[
\phi_{\mathrm{out}}(x)
=
\Phi_{\mathrm{out}}
(M_{\mathrm{ft}},M_{\mathrm{pre}},x).
\]

\paragraph{Router-telemetry features.}
We summarize the per-token routing information into a fixed-dimensional
representation for each input. Specifically, we use the average routing
probability assigned to each expert. For expert $e$ at MoE layer $\ell$,
we define
\[
r_{\ell,e}(x)
=
\frac{1}{T}
\sum_{t=1}^{T}
p_{\ell,t,e},
\]
where $p_{\ell,t,e}$ is the routing probability assigned to expert $e$
for token $t$. Concatenating these statistics across all MoE layers and
experts gives the router representation
\[
\phi_R(x)
=
\left[
r_{\ell,e}(x)
\right]_{\ell,e}.
\]

\paragraph{Router-augmented membership classifier.}
Since the victim's private fine-tuning data are unknown, the attacker
constructs shadow models to learn membership patterns applicable to the
target model. For each $j\in\{1,\ldots,K\}$, the attacker constructs a
fine-tuning set $\widetilde{D}^{\mathrm{in}}_j$ and a held-out set
$\widetilde{D}^{\mathrm{out}}_j$ from auxiliary data in the same
domain, and fine-tunes $M_{\mathrm{pre}}$ on
$\widetilde{D}^{\mathrm{in}}_j$ to obtain a shadow model
$\widetilde{M}_j$. For the $j$-th shadow model, we define
\[
\widetilde{m}_j(x)
=
\begin{cases}
1, & x\in\widetilde{D}^{\mathrm{in}}_j,\\
0, & x\in\widetilde{D}^{\mathrm{out}}_j.
\end{cases}
\]

For each shadow model $\widetilde{M}_j$, the attacker extracts the
corresponding output-side and router representations
$\phi_{\mathrm{out}}^{(j)}(x)$ and $\phi_R^{(j)}(x)$. The
representations from all $K$ shadow models are pooled to form the
attack-training dataset
\[
\mathcal{D}_{\mathrm{attack}}
=
\bigcup_{j=1}^{K}
\left\{
\left(
[\phi_{\mathrm{out}}^{(j)}(x);
 \phi_R^{(j)}(x)],
\widetilde{m}_j(x)
\right)
:
x\in
\widetilde{D}^{\mathrm{in}}_j
\cup
\widetilde{D}^{\mathrm{out}}_j
\right\}.
\]

The attacker trains a membership classifier $g$ on
$\mathcal{D}_{\mathrm{attack}}$. For a candidate example $x$ queried
against the target model, the membership score is
\[
s(x)
=
g\!\left(
[\phi_{\mathrm{out}}(x);\phi_R(x)]
\right).
\]
The classifier learned from the shadow models is then applied directly
to the independently fine-tuned target model $M_{\mathrm{ft}}$.

\section{Experiments}
\label{sec:exp}

In this section, we empirically evaluate the additional membership leakage introduced by exposed router telemetry.
We first examine whether router telemetry provides additional leakage beyond output-side membership signals, and whether this effect persists across architectures, domains, adaptation regimes, and restricted attacker access.
We then examine whether the leakage is specific to the router or instead reflects membership information encoded more broadly in the hidden representations during fine-tuning.
Finally, we evaluate telemetry perturbations as potential mitigations and examine the resulting privacy--utility trade-off.

\subsection{Experimental Setup}
\label{sec:exp_setup}

\textbf{Models and datasets.}
We consider three open-weight MoE models: OLMoE-1B-7B-0924 (OLMoE-1B-7B)~\citep{OLMoE}, Qwen1.5-MoE-A2.7B-Chat (Qwen1.5-MoE)~\citep{qwen_moe}, and LLaMA-MoE-v2-3.8B-2.8-sft (LLaMA-MoE-v2)~\citep{llama-moe-v2}.
We evaluate each model on three domains: arXiv, GitHub, and Wikipedia. 
More details are provided in Appendix~\ref{app:data}.

\textbf{Shadow-model construction.}
For each model--domain pair, we partition the corresponding domain corpus into four mutually disjoint subsets. 
For each subset $j$, we construct a fine-tuning set $\widetilde{D}^{\mathrm{in}}_j$ and a held-out set $\widetilde{D}^{\mathrm{out}}_j$ (with text-only blind classifiers~\citep{das2025blind} remaining near chance in distinguishing the two sets).
We then fine-tune the corresponding pretrained model on $\widetilde{D}^{\mathrm{in}}_j$ to obtain an independently fine-tuned model $\widetilde{M}_j$.
Full construction, validation, and fine-tuning details are provided in Appendix~\ref{app:finetuning}.

\textbf{MIA baselines.}
We consider a set of output-side membership inference signals, including NLL~\citep{MIA_NLL}, Min-K\%~\citep{MIA_Min-K}, Min-K\%++~\citep{MIA_Min-K++}, zlib-normalized likelihood~\citep{MIA_Zlib}, confidence-derived logit margin~\citep{MIA_Confidence}, log-probability percentile statistics~\citep{MIA_Min-K,MIA_SURP}, reference loss drop~\citep{MIA_Reference-loss,MIA_LiRA}, WBC~\citep{MIA_WBC}, EZ~\citep{MIA_EZ}, ReCaLL~\citep{MIA_ReCaLL}, and LTsig~\citep{MIA_LTsig}.
We evaluate these signals individually and also concatenate them into an ensemble output-side representation, denoted by $A_{\mathrm{out}}$, which is used by the membership classifier.
Detailed definitions of the individual signals are provided in Appendix~\ref{app:baselines}.

\textbf{Evaluation protocol.}
For each model--domain pair, we perform leave-one-out evaluation over the four independently fine-tuned models: one is treated as the target and the remaining three as shadow models. 
Both the output-only and router-augmented classifiers are trained on the pooled shadow-model data and applied directly to the held-out target, with the latter additionally incorporating router telemetry. 
We repeat this procedure over all four held-out targets and pool the resulting predictions for
evaluation. 
We use histogram gradient boosting as the membership classifier and continuous expert-load telemetry as the main router signal.

\subsection{Router Telemetry Exposure Amplifies Membership Leakage}
\label{sec:main_results}

We first examine whether exposing router telemetry increases the membership inference risk of fine-tuned MoE language models.
To this end, we evaluate router-augmented membership inference across three MoE architectures and three data domains, covering all nine model--domain combinations.
As shown in Table~\ref{tab:main_grid}, augmenting the output-signal ensemble $A_{\mathrm{out}}$ with router telemetry consistently improves $\operatorname{TPR}@1\%\operatorname{FPR}$ in all nine settings, with absolute gains ranging from 0.0267 to 0.0937.
These results show that exposing router telemetry introduces additional membership leakage beyond that already captured by conventional output-side signals, and that this effect generalizes across different MoE architectures and data domains.

Router telemetry also improves AUROC across all nine settings, although the
gains are smaller because $A_{\mathrm{out}}$ is already close to saturation
in several cases.
In comparison, router telemetry yields more pronounced gains at the stricter
low-FPR operating point measured by
$\operatorname{TPR}@1\%\operatorname{FPR}$.
The same trend also holds for individual output-side MIA baselines, with
detailed per-baseline results reported in
Appendix~\ref{app:baseline_results}.

\begin{table*}[t]
\centering
\caption{
Membership inference across all combinations of three MoE architectures and three data domains.
$\Delta$ denotes the absolute improvement after incorporating router telemetry into the output-signal ensemble $A_{\mathrm{out}}$.
}
\label{tab:main_grid}
\small
\begin{tabular}{llrrrrrr}
\toprule
Model
& Domain
& \multicolumn{3}{c}{TPR@1\% FPR}
& \multicolumn{3}{c}{AUROC} \\
\cmidrule(lr){3-5}
\cmidrule(lr){6-8}
& & $A_{\mathrm{out}}$ & +Router & $\Delta$
& $A_{\mathrm{out}}$ & +Router & $\Delta$ \\
\midrule
OLMoE-1B-7B
& arXiv
& 0.8686 & 0.9527 & +0.0841
& 0.9913 & 0.9974 & +0.0061 \\
OLMoE-1B-7B
& GitHub
& 0.8503 & 0.9133 & +0.0630
& 0.9934 & 0.9960 & +0.0025 \\
OLMoE-1B-7B
& Wikipedia
& 0.9388 & 0.9655 & +0.0267
& 0.9977 & 0.9986 & +0.0009 \\
\midrule
Qwen1.5-MoE
& arXiv
& 0.8610 & 0.9343 & +0.0733
& 0.9909 & 0.9965 & +0.0056 \\
Qwen1.5-MoE
& GitHub
& 0.7858 & 0.8615 & +0.0757
& 0.9905 & 0.9939 & +0.0033 \\
Qwen1.5-MoE
& Wikipedia
& 0.9068 & 0.9475 & +0.0407
& 0.9954 & 0.9976 & +0.0022 \\
\midrule
LLaMA-MoE-v2
& arXiv
& 0.8775 & 0.9220 & +0.0445
& 0.9926 & 0.9952 & +0.0026 \\
LLaMA-MoE-v2
& GitHub
& 0.6725 & 0.7662 & +0.0937
& 0.9832 & 0.9887 & +0.0054 \\
LLaMA-MoE-v2
& Wikipedia
& 0.8278 & 0.8927 & +0.0648
& 0.9911 & 0.9941 & +0.0030 \\
\bottomrule
\end{tabular}
\end{table*}

\subsection{Leakage Persists Across Fine-Tuning Regimes}
\label{sec:adaptation}

\begin{table}[t]
\centering
\caption{
Membership inference under different fine-tuning regimes on OLMoE-1B-7B with arXiv, measured by $\operatorname{TPR}@1\%\operatorname{FPR}$.
$\Delta$ denotes the improvement after incorporating router telemetry.
}
\label{tab:adaptation}
\small
\begin{tabular}{lccc}
\toprule
Fine-tuning regime
& $A_{\mathrm{out}}$
& +Router 
& $\Delta$ \\
\midrule
Full fine-tuning
& 0.8686 & 0.9527 & +0.0841 \\
Frozen router
& 0.8612 & 0.9516 & +0.0904 \\
LoRA
& 0.9226 & 0.9531 & +0.0305 \\
Instruction tuning
& 0.8618 & 0.9300 & +0.0682 \\
\bottomrule
\end{tabular}
\end{table}

Having established that router telemetry provides additional membership signal beyond model outputs, we next ask whether this effect depends on the fine-tuning procedure.
Using OLMoE-1B-7B on arXiv, we compare standard full fine-tuning with frozen-router fine-tuning, LoRA adaptation~\citep{hu2021loralowrankadaptationlarge}, and instruction tuning~\citep{wei2022finetuned} with response-only loss.
As shown in Table~\ref{tab:adaptation}, incorporating router telemetry improves $\operatorname{TPR}@1\%\operatorname{FPR}$ under all four regimes, with gains ranging from 0.0305 under LoRA to 0.0904 with a frozen router.

The leakage therefore persists across substantially different adaptation settings.
In particular, freezing the router does not reduce the additional leakage, with a gain comparable to that under full fine-tuning.
The effect also remains under LoRA and instruction tuning, showing that it is not specific to full-parameter adaptation.
More detailed settings and results are provided in Appendix~\ref{app:adaptation}.

\subsection{Leakage Under Restricted Telemetry and Shadow-Model Access}
\label{sec:access_requirements}

The experimental results in Section~\ref{sec:main_results} assume access to the full continuous router-telemetry representation.
We next examine whether the observed membership leakage remains exploitable under more restricted access conditions that may arise in practical deployment settings.
Specifically, we consider three settings: replacing continuous routing probabilities with discrete expert selections, restricting the amount of continuous telemetry exposed to the attacker, and reducing the number of shadow models available for attack training.

\textbf{Discrete expert-selection telemetry.}
The experimental results in Section~\ref{sec:main_results} use the continuous expert-load representation
$\phi_R$, which retains the routing probabilities
assigned to individual experts.
We first consider a more restricted representation that exposes only
which experts are selected by the router.
For each layer $\ell$ and expert $e$, we define
\begin{equation}
    r^{\mathrm{count}}_{\ell,e}(x)
    =
    \frac{1}{T}
    \sum_{t=1}^{T}
    \mathbb{I}
    \left[
        e \in \mathcal{E}_{\ell,t}^{(k)}
    \right],
\end{equation}
where $\mathcal{E}_{\ell,t}^{(k)}$ denotes the set of experts selected
for token $t$ at layer $\ell$.
Flattening these layer--expert selection frequencies gives the discrete
representation $\phi_R^{\mathrm{count}}(x)$, which contains no
continuous routing probabilities.

As shown in Table~\ref{tab:expert_id_results}, augmenting $A_{\mathrm{out}}$ with the discrete expert-selection representation $\phi_R^{\mathrm{count}}$ improves $\operatorname{TPR}@1\%\operatorname{FPR}$ in all nine model--domain settings, with gains ranging from 0.0283 to 0.1043.
These improvements are comparable in magnitude to those obtained with the continuous router representation $\phi_R$, and in several settings the discrete representation achieves a similar or slightly larger gain.
This indicates that the additional membership leakage does not depend on access to precise routing probabilities, as discrete expert-selection information alone is sufficient to expose a comparable membership signal.

\begin{table}[t]
\centering
\caption{
Membership inference with discrete expert-selection telemetry.
$\Delta(\phi_R^{\mathrm{count}})$ denotes the improvement in
$\operatorname{TPR}@1\%\operatorname{FPR}$ obtained by incorporating
the discrete expert-selection representation, while
$\Delta(\phi_R)$ reports the corresponding improvement
with the continuous expert-load representation.
}
\label{tab:expert_id_results}
\small
\begin{tabular}{llcccc}
\toprule
Model & Domain
& $A_{\mathrm{out}}$
& +Router $(\phi_R^{\mathrm{count}})$
& $\Delta(\phi_R^{\mathrm{count}})$
& $\Delta(\phi_R)$ \\
\midrule
OLMoE-1B-7B
& arXiv
& 0.8686
& 0.9400
& $+0.0714$
& $+0.0841$ \\
OLMoE-1B-7B
& GitHub
& 0.8503
& 0.9163
& $+0.0660$
& $+0.0630$ \\
OLMoE-1B-7B
& Wikipedia
& 0.9388
& 0.9671
& $+0.0283$
& $+0.0267$ \\
\midrule
Qwen1.5-MoE
& arXiv
& 0.8610
& 0.9241
& $+0.0631$
& $+0.0733$ \\
Qwen1.5-MoE
& GitHub
& 0.7858
& 0.8730
& $+0.0872$
& $+0.0757$ \\
Qwen1.5-MoE
& Wikipedia
& 0.9068
& 0.9453
& $+0.0385$
& $+0.0407$ \\
\midrule
LLaMA-MoE-v2
& arXiv
& 0.8775
& 0.9114
& $+0.0339$
& $+0.0445$ \\
LLaMA-MoE-v2
& GitHub
& 0.6725
& 0.7768
& $+0.1043$
& $+0.0937$ \\
LLaMA-MoE-v2
& Wikipedia
& 0.8278
& 0.8870
& $+0.0592$
& $+0.0648$ \\
\bottomrule
\end{tabular}
\end{table}

\textbf{Restricted continuous telemetry.}
We next ask how the leakage changes when only a subset of the
continuous expert-load telemetry is exposed.
Using OLMoE-1B-7B on arXiv, we retain only the top-$K$ experts ranked
by document-level routing load to restrict the amount of continuous
router telemetry exposed to the attacker.

\begin{wrapfigure}{r}{0.42\textwidth}
    \centering
    \vspace{-8pt}
    \includegraphics[width=\linewidth]{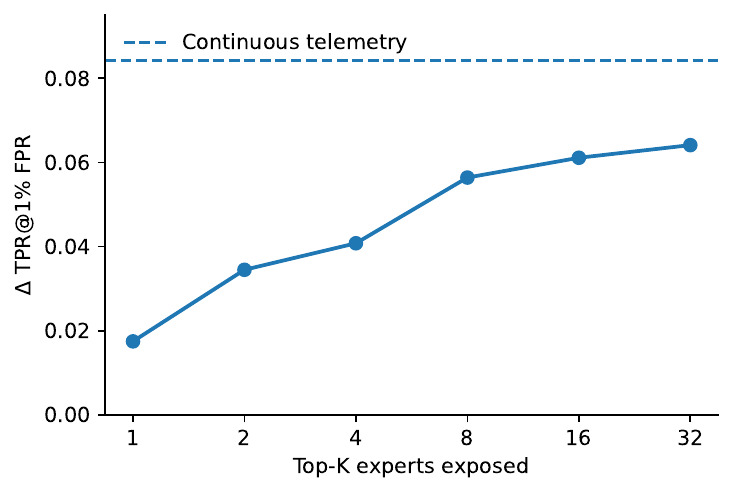}
    \caption{
    Effect of restricted continuous router telemetry on OLMoE-1B-7B with arXiv.
    The dashed line denotes the full continuous telemetry setting.
    }
    \label{fig:topk_tpr}
    \vspace{-8pt}
\end{wrapfigure}

As shown in Figure~\ref{fig:topk_tpr}, even top-1 telemetry provides an additional $\Delta\operatorname{TPR}@1\%\operatorname{FPR}$ of 0.0175 over $A_{\mathrm{out}}$.
The gain increases monotonically with top-$K$, reaching 0.0564 at top-$K=8$ and 0.0641 at top-$K=32$, with diminishing improvements as more experts are exposed.
Thus, membership leakage remains observable even when only a small fraction of the continuous routing information is available.

\textbf{Limited shadow-model access.}
Finally, we consider limited shadow-model access.
Using only a single shadow model still improves
$\operatorname{TPR}@1\%\operatorname{FPR}$ over $A_{\mathrm{out}}$ by
0.0493, showing that the router-induced membership signal remains
exploitable even with minimal shadow-model resources.
Additional results are reported in Appendix~\ref{app:shadow_count}.

\subsection{Router Telemetry Exposes Membership Signals in Hidden Representations}
\label{sec:mechanism}

\begin{wrapfigure}{r}{0.42\textwidth}
    \centering
    \vspace{-10pt}
    \includegraphics[width=\linewidth]{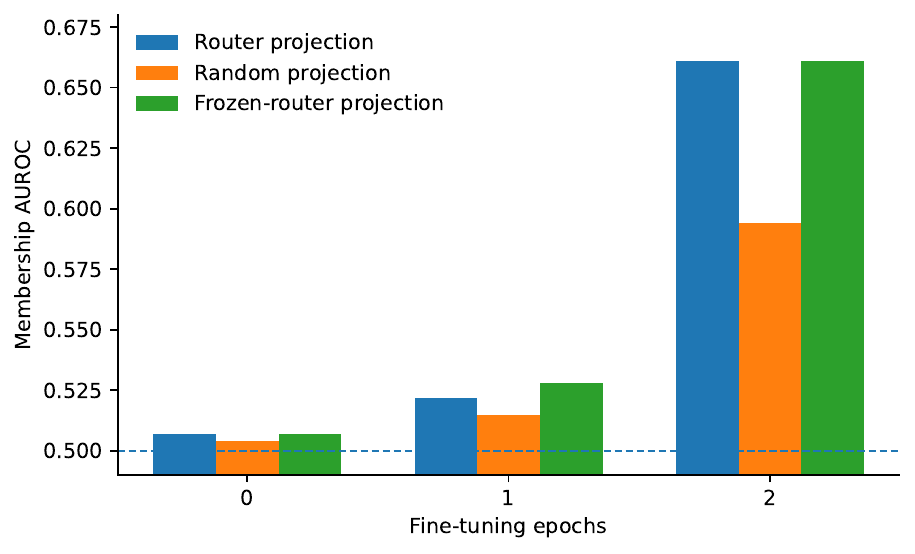}
    \caption{
    Mechanistic analysis on OLMoE-1B-7B with arXiv.
    Membership AUROC of matched-rank random, router, and frozen-router
    projections across fine-tuning epochs.
    The dashed line denotes chance-level AUROC of 0.5.
    }
    \label{fig:mechanism}
    \vspace{-10pt}
\end{wrapfigure}

Having established that router telemetry amplifies membership leakage, we
next investigate whether this leakage is specific to the router or instead
reflects membership information encoded more broadly in the hidden
representations during fine-tuning. Using OLMoE-1B-7B on arXiv, we compare
the router projection with a matched-rank random projection under full
fine-tuning and with the corresponding router projection under frozen-router
fine-tuning to examine whether the observed membership signal is specific
to the router subspace or depends on router-parameter updates.

As shown in Figure~\ref{fig:mechanism}, all three conditions are near the chance-level AUROC of 0.5 before fine-tuning. 
After one epoch, the router and random projections reach AUROCs of 0.522 and 0.515, respectively, increasing to 0.661 and 0.594 after two epochs. 
The router projection therefore retains more membership information than a matched-rank random
subspace. 
Freezing the router produces a similar trend, with the router projection reaching  an AUROC of 0.661 after two epochs under both full and frozen-router fine-tuning.

Together, these results suggest that the observed leakage does not require router-specific memorization. 
Instead, fine-tuning introduces membership-related information into the hidden representations, and router telemetry exposes a projection of this signal during inference.

\subsection{Mitigating Router-Telemetry Leakage Trades Off Utility}
\label{sec:defense}

Having established that exposed router telemetry introduces additional
membership leakage, we next examine whether this risk can be mitigated by
perturbing the telemetry before it is exposed.
Using OLMoE-1B-7B on arXiv, we consider three post-processing mechanisms
applied to the continuous expert-load representation: Gaussian noise,
random feature dropout, and quantization.
These perturbations are applied after inference and do not modify the
model or its inference procedure.
We measure the remaining leakage by the improvement in
$\operatorname{TPR}@1\%\operatorname{FPR}$ over $A_{\mathrm{out}}$,
and telemetry utility by its Pearson correlation with the original expert
loads.
We use correlation as the primary utility metric in the main analysis,
while top-1 expert agreement, which shows the same qualitative trend, is
reported in Appendix~\ref{app:defense}.

As shown in Figure~\ref{fig:defense}, all three perturbation mechanisms reduce the additional membership leakage as the perturbation becomes stronger, while simultaneously degrading telemetry utility. 
For Gaussian noise, the $\operatorname{TPR}@1\%\operatorname{FPR}$ gain decreases from 0.0841 without perturbation to 0.0563 at $\sigma=1$ and 0.0159 at $\sigma=4$, while telemetry correlation decreases from 1.0 to 0.924 and 0.588, respectively. 
Feature dropout produces a stronger reduction in both quantities, with the gain and correlation decreasing to 0.0190 and 0.153 at $p=0.9$, and to 0.0040 and 0.047 at $p=0.99$.
Similarly, quantization reduces the gain from 0.0719 at $B=32$ to 0.0121 at $B=8$ and 0.0079 at $B=4$, while correlation decreases from 0.750 to 0.584 and 0.045.

Overall, stronger telemetry perturbation can substantially suppress the additional membership leakage, but doing so also reduces the fidelity of the exposed telemetry. 
In particular, the strongest suppression under dropout and quantization occurs only when telemetry correlation has already fallen close to zero. 
Within the evaluated settings, we do not observe a configuration that suppresses the router-induced leakage while preserving high-fidelity telemetry.

\begin{figure*}[t]
    \centering

    \begin{subfigure}[t]{0.32\textwidth}
        \centering
        \includegraphics[width=\linewidth]
        {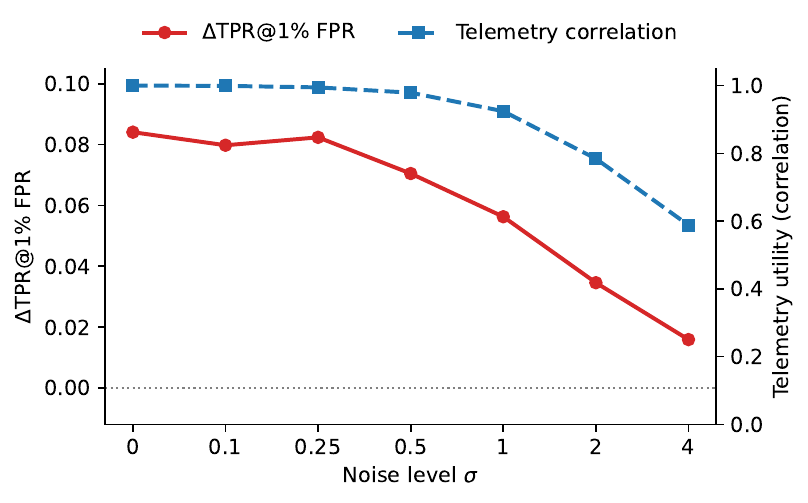}
        \caption{Gaussian noise.}
        \label{fig:defense_gaussian}
    \end{subfigure}
    \hfill
    \begin{subfigure}[t]{0.32\textwidth}
        \centering
        \includegraphics[width=\linewidth]
        {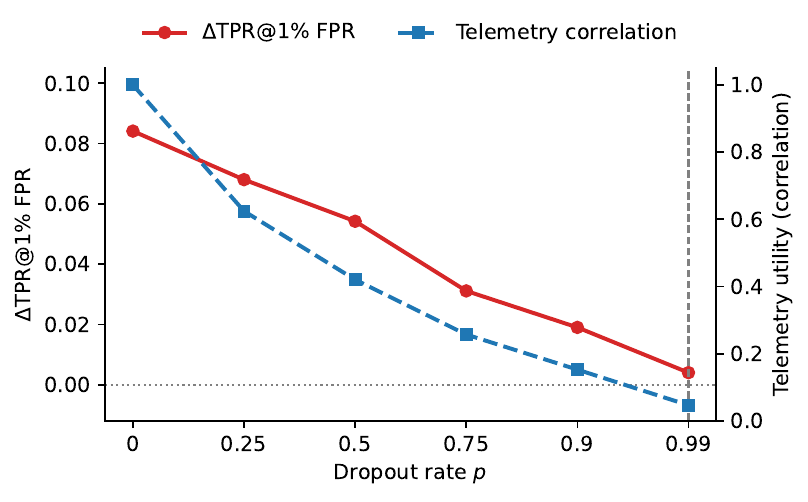}
        \caption{Feature dropout.}
        \label{fig:defense_dropout}
    \end{subfigure}
    \hfill
    \begin{subfigure}[t]{0.32\textwidth}
        \centering
        \includegraphics[width=\linewidth]
        {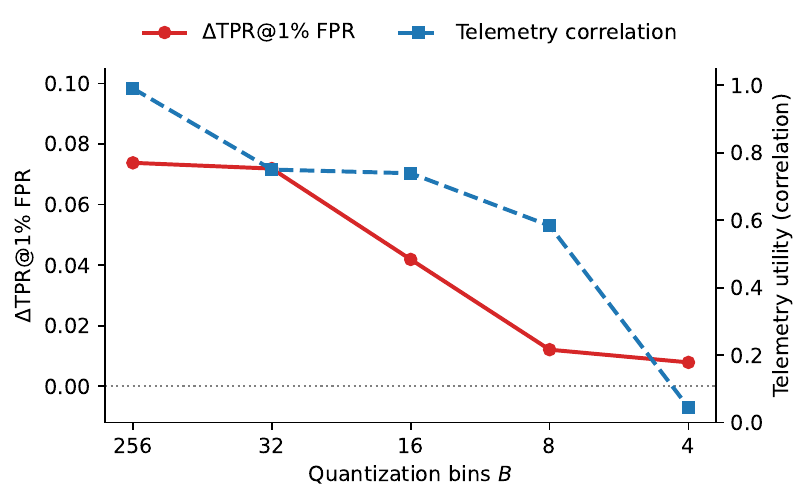}
        \caption{Quantization.}
        \label{fig:defense_quantization}
    \end{subfigure}

    \caption{
    Router-telemetry perturbation on OLMoE-1B-7B with arXiv.
    In each panel, the left y-axis reports the improvement in
    $\operatorname{TPR}@1\%\operatorname{FPR}$ over
    $A_{\mathrm{out}}$, and the right y-axis reports telemetry utility
    measured by Pearson correlation with the original expert loads.
    The x-axis shows the corresponding perturbation parameter, with
    stronger perturbation given by larger $\sigma$ or $p$ and smaller $B$.
    }
    \label{fig:defense}
\end{figure*}

\section{Conclusion}

We study membership-inference risk introduced by router telemetry in fine-tuned Mixture-of-Experts language models. Our results show that routing information exposed during inference provides membership signal beyond that available from conventional output-side attacks across multiple MoE architectures, data domains, and fine-tuning regimes. This additional leakage does not require access to precise routing probabilities: discrete expert selections, restricted top-$K$ telemetry, and even limited shadow-model access remain sufficient to improve membership inference. Our mechanistic analysis further shows that the effect is not explained by router-specific memorization. Instead, fine-tuning induces membership-related structure in hidden representations, and the router exposes a projection of this information even when its parameters are frozen during fine-tuning. Finally, post-processing router telemetry can reduce the additional leakage, but stronger perturbations also degrade the fidelity of the telemetry itself. Taken together, these findings show that router telemetry should not be treated as an innocuous operational signal: exposing routing information can reveal properties of a model's private fine-tuning data and therefore introduces an additional privacy consideration when deploying MoE language models.

\section*{AI Use Statement}

We used generative AI tools to assist with polishing the writing of this manuscript, including improving grammar, clarity, and readability.
All AI-assisted text was reviewed and revised by the authors.
Generative AI was not used to generate research ideas, experimental results, or scientific claims.
The authors take full responsibility for the final content of this work.

\section*{Ethics Statement}

This work studies a privacy risk associated with exposing router telemetry from fine-tuned Mixture-of-Experts language models.
As with other membership inference research, the techniques studied in this paper have potential dual-use implications, since they could be used to infer whether particular examples were included in a model's fine-tuning data.
Our experiments are conducted in controlled settings using open-weight models and constructed membership instances from arXiv, GitHub, and Wikipedia, and do not involve human subjects or access to private user data from deployed systems.
The purpose of this work is to characterize an underexplored privacy surface in MoE deployment and to inform safer telemetry-exposure practices.
We also evaluate several telemetry-perturbation mechanisms to quantify the trade-off between reducing membership leakage and preserving the operational utility of the exposed telemetry.
We hope these findings encourage model providers to consider routing information as potentially privacy-sensitive when designing logging, monitoring, and telemetry interfaces.

\bibliography{iclr2027_conference}
\bibliographystyle{iclr2027_conference}

\clearpage
\appendix

\section{Experimental Setup}
\label{app:setup}

\subsection{Dataset Construction and Validation}
\label{app:data}

We evaluate membership inference on three domains: arXiv, GitHub, and Wikipedia. 
For each domain, we construct four independent membership instances from a common source pool. 
Each instance $j\in\{1,\ldots,4\}$ consists of a fine-tuning set $\widetilde{D}^{\mathrm{in}}_j$ and a held-out set $\widetilde{D}^{\mathrm{out}}_j$, corresponding to member and non-member examples, respectively. 
The source corpus is shuffled before construction, and candidate documents are assigned so that no document is shared across different instances:
\begin{equation}
\left(
\widetilde{D}^{\mathrm{in}}_j
\cup
\widetilde{D}^{\mathrm{out}}_j
\right)
\cap
\left(
\widetilde{D}^{\mathrm{in}}_{j'}
\cup
\widetilde{D}^{\mathrm{out}}_{j'}
\right)
=
\varnothing,
\qquad j\neq j'.
\label{eq:disjoint_instances}
\end{equation}
This prevents the membership classifier from benefiting from repeated documents across shadow and target models.

For arXiv, each instance contains 2,500 members and 2,500 non-members, drawn from a source pool of 24,606 documents. 
For GitHub and Wikipedia, each instance contains 1,500 members and 1,500 non-members, with a source pool of 12,000 documents per domain. 
The same four data partitions are reused across all three model architectures within a domain. 
Thus, when comparing architectures, the underlying membership instances are held fixed and only the model changes. Dataset statistics are summarized in Table~\ref{tab:dataset_statistics}.

\begin{table}[t]
\centering
\caption{
Dataset statistics used for the main experiments.
$N_{\mathrm{in}}$ and $N_{\mathrm{out}}$ denote the number of member
and non-member examples, respectively, in each of the four independent
membership instances.
}
\label{tab:dataset_statistics}
\small
\begin{tabular}{lrrr}
\toprule
Domain
& Source pool
& $N_{\mathrm{in}}$
& $N_{\mathrm{out}}$ \\
\midrule
arXiv      & 24,606 & 2,500 & 2,500 \\
GitHub     & 12,000 & 1,500 & 1,500 \\
Wikipedia  & 12,000 & 1,500 & 1,500 \\
\bottomrule
\end{tabular}
\end{table}

We additionally verify that the member and non-member partitions do not contain readily exploitable surface-text differences. 
Such differences can allow model-free "blind" classifiers to outperform membership inference attacks, making attack performance difficult to interpret~\citep{das2025blind}. 
Following this evaluation practice, for each of the 12 domain–instance pairs, we train four text-only classifiers that use no information from either the pretrained or fine-tuned model: word TF--IDF features with 1--2-grams, character TF--IDF features with 3--5-grams, bag-of-words features, and hand-crafted surface statistics.
Each classifier is evaluated using five-fold out-of-fold prediction.
Across all 48 checks, AUROC ranges from 0.4817 to 0.5329, with no systematic deviation from chance. This sanity check indicates that the membership labels are not readily predictable from surface properties of the constructed datasets alone.

\subsection{Models and Fine-Tuning}
\label{app:finetuning}

We consider three open-weight MoE language models: OLMoE-1B-7B-0924~\citep{OLMoE}, Qwen1.5-MoE-A2.7B-Chat~\citep{qwen_moe}, and LLaMA-MoE-v2-3.8B-2.8-sft~\citep{llama-moe-v2}.
Combining the three architectures with the three domains above yields the $3\times3$ model--domain grid used in the main experiments.

For each model--domain pair and each membership instance $j$, we fine-tune the corresponding pretrained model on $\widetilde{D}^{\mathrm{in}}_j$ to obtain an independently adapted model $\widetilde{M}_j$. Unless otherwise specified, the main experiments use full-parameter fine-tuning, including the router parameters. 
We use Adafactor with a learning rate of $8\times10^{-6}$, a training batch size of 16, and a maximum sequence length of 256 tokens. 
All main results are evaluated after one epoch of fine-tuning, and the same configuration is used across all nine model--domain settings.

The four models within each model--domain setting are trained independently on mutually disjoint fine-tuning sets. 
In the main leave-one-out evaluation, each model is in turn treated as the target model while the remaining three act as shadow models.

\subsection{MIA Baseline Implementation}
\label{app:baselines}

We evaluate output-side membership signals derived from the fine-tuned
target model and, when required, the public reference checkpoint.
Unless otherwise specified, the resulting feature representations are
provided to the same histogram gradient boosting classifier used
throughout our experiments.

\paragraph{Reference-free likelihood and confidence features.}
We first construct a 13-dimensional feature block, denoted
$\phi_{\mathrm{strong}}$, from token-level statistics of the fine-tuned
model. Let $\ell_t$ denote the log-probability assigned to the true token
at position $t$.

\textbf{NLL.}
We use the mean negative log-likelihood over all valid tokens as a
one-dimensional feature.

\textbf{Min-K\%.}
We sort the token log-probabilities in ascending order and compute the
mean over the lowest $K\%$ of tokens for
$K\in\{10,20,30\}$.
The resulting three values form a three-dimensional Min-K\% feature.

\textbf{Min-K\%++.}
For each token, we normalize the target-token logit using the mean and
standard deviation of the full vocabulary-logit distribution at that
position:
\begin{equation}
    z_t
    =
    \frac{u_t-\mu_t}{\sigma_t},
\end{equation}
where $u_t$ is the true-token logit and $\mu_t$ and $\sigma_t$ are the
mean and standard deviation of the vocabulary logits.
We sort $\{z_t\}$ in ascending order and average the lowest
$K\%$ for $K\in\{10,20\}$, yielding a two-dimensional feature.

\textbf{zlib-normalized likelihood.}
Let
\begin{equation}
    \rho_{\mathrm{zlib}}(x)
    =
    \frac{
        |\operatorname{zlib}(x)|
    }{
        |x|
    }
\end{equation}
denote the compression ratio of the UTF-8 representation of the
document.
We use both $\rho_{\mathrm{zlib}}(x)$ and
\begin{equation}
    \frac{\operatorname{NLL}(x)}
         {\rho_{\mathrm{zlib}}(x)+\epsilon}
\end{equation}
as features, giving a two-dimensional zlib block.

\textbf{Logit margin.}
At each valid token position, we compute the difference between the
largest and second-largest vocabulary logits and average this quantity
over the document, yielding a single scalar feature.

\textbf{Log-probability percentile statistics.}
From the sequence of true-token log-probabilities, we compute the
10th, 50th, and 90th percentiles together with the standard deviation.
These four statistics form a four-dimensional feature block.

The six feature groups above contain 13 dimensions in total and
constitute the reference-free feature block used as part of
$A_{\mathrm{out}}$.

\paragraph{Reference loss drop.}
Let $\ell_t^{\mathrm{ft}}$ and $\ell_t^{\mathrm{pre}}$ denote the
true-token log-probabilities under the fine-tuned and reference models,
respectively. We compute
\begin{equation}
    \phi_{\mathrm{refdrop}}(x)
    =
    \frac{1}{T}
    \sum_{t=1}^{T}
    \left(
        \ell_t^{\mathrm{ft}}
        -
        \ell_t^{\mathrm{pre}}
    \right),
\end{equation}
using the positions available under both models.

\paragraph{WBC.}
Following the window-based comparison between the fine-tuned and
reference models, we consider ten window sizes
\begin{equation}
    \mathcal{W}
    =
    \{2,3,4,5,8,11,15,21,29,40\}.
\end{equation}
For each window size $w$, we slide a window across the document and
compute the fraction of windows for which the summed log-probability
under the fine-tuned model exceeds that under the reference model.
We denote this vote fraction by $v_w$ and compute
\begin{equation}
    S_{\mathrm{WBC}}
    =
    \frac{1}{|\mathcal{W}|}
    \sum_{w\in\mathcal{W}} v_w.
\end{equation}
The individual WBC baseline uses only $S_{\mathrm{WBC}}$.
For $A_{\mathrm{out}}$, we additionally retain the vote fractions
$\{v_w\}_{w\in\mathcal{W}}$.

\paragraph{EZ.}
Let $\mathcal{E}$ denote positions where the fine-tuned model's
top-1 prediction differs from the ground-truth token, and define
\begin{equation}
    d_t
    =
    \ell_t^{\mathrm{ft}}
    -
    \ell_t^{\mathrm{pre}},
    \qquad t\in\mathcal{E}.
\end{equation}
We compute
\begin{equation}
    P=\sum_{t\in\mathcal{E}}\max(d_t,0),
    \qquad
    N=\sum_{t\in\mathcal{E}}\max(-d_t,0).
\end{equation}
The individual EZ baseline uses the ratio
\begin{equation}
    S_{\mathrm{EZ}}
    =
    \frac{P}{N+\epsilon}.
\end{equation}
Within $A_{\mathrm{out}}$, we use the four-dimensional block
\begin{equation}
    \left[
        \frac{P}{N+\epsilon},
        \log(1+P),
        \log(1+N),
        |\mathcal{E}|/T
    \right].
\end{equation}

\paragraph{ReCaLL.}
We use a fixed prefix constructed from eight held-out non-member documents. 
The concatenated prefix is tokenized and truncated to at most 128 tokens and is reused for all evaluated candidates. 
For candidate $x$, ReCaLL is computed as the difference between its mean token log-probability under the fine-tuned model when conditioned on this prefix and its mean token log-probability without the prefix. 
The original ReCaLL score is defined as the ratio of the conditional to the unconditional log-likelihood~\citep{MIA_ReCaLL}; we use their difference as a simpler variant that captures the same prefix-induced change. 
This signal uses only the fine-tuned model and does not require the reference checkpoint.

\paragraph{LTsig.}
We construct a signature from token-level differences between the
fine-tuned and reference models.
Let
\begin{equation}
    d_t
    =
    \ell_t^{\mathrm{ft}}
    -
    \ell_t^{\mathrm{pre}}.
\end{equation}
The signature contains 16 uniformly spaced quantiles of $\{d_t\}$,
eight uniformly spaced quantiles of
$\{\ell_t^{\mathrm{ft}}\}$, and eight summary statistics:
\begin{equation}
\begin{split}
    [&
    \operatorname{mean}(d),
    \operatorname{std}(d),
    \Pr(d_t>0),
    \sum_t d_t, \\
    &
    \operatorname{mean}(r^{\mathrm{ft}}),
    \operatorname{mean}(r^{\mathrm{pre}}),
    \operatorname{mean}(r^{\mathrm{ft}}-r^{\mathrm{pre}}),
    T
    ] ,
\end{split}
\end{equation}
where $r^{\mathrm{ft}}$ and $r^{\mathrm{pre}}$ denote the corresponding
token-rank statistics.
This produces a 32-dimensional feature representation.
Our implementation follows the token-level target--reference signature
idea of LTsig, but uses this fixed-dimensional summary with the common
histogram gradient boosting classifier rather than reproducing the
original sequence-based classifier.

\paragraph{Output-signal ensemble.}
Our primary output-side baseline, denoted by $A_{\mathrm{out}}$,
concatenates the 13-dimensional likelihood/confidence block with
reference loss drop, the full WBC feature block, the full EZ feature
block, ReCaLL, and the 32-dimensional LTsig-style signature:
\begin{equation}
    \phi_{\mathrm{out}}(x)
    =
    \operatorname{concat}_{b\in\mathcal{B}}
    \phi_b(x).
\end{equation}
A histogram gradient boosting classifier is then trained on
$\phi_{\mathrm{out}}(x)$.

For the individual-baseline analysis in Appendix~\ref{app:baseline_results}, we evaluate each feature family separately.
The standalone WBC and EZ rows use $S_{\mathrm{WBC}}$ and $S_{\mathrm{EZ}}$, respectively, while $A_{\mathrm{out}}$ retains their full feature blocks as described above.

\subsection{Router-Telemetry Representation}
\label{app:telemetry}

Our primary telemetry representation is the continuous expert-load feature introduced in Section~\ref{sec:method}. 
For each MoE layer $\ell$ and expert $e$, we average the softmax routing probability across all tokens in the candidate sequence:
\begin{equation}
r_{\ell,e}(x)
=
\frac{1}{T}
\sum_{t=1}^{T}
p_{\ell,t,e}.
\label{eq:appendix_router_load}
\end{equation}
We then flatten the per-layer expert loads into
\begin{equation}
\phi_R(x)
=
\left[
r_{\ell,e}(x)
\right]_{\ell,e}.
\label{eq:router_load_feature}
\end{equation}
This continuous representation is used by default throughout the main experiments.

We separately evaluate more restricted forms of telemetry. 
In particular, $\phi_R^{\mathrm{count}}$ records the frequency with which each expert is selected by the router rather than its continuous routing probability, thereby retaining only discrete expert-selection information. 

\subsection{Attack Training and Evaluation}
\label{app:evaluation}

We use histogram gradient boosting as the default membership classifier.
Both output-only and router-augmented attacks follow the same leave-one-out shadow-model evaluation protocol.

\paragraph{Leave-one-out shadow-model evaluation.}
For each model--domain setting, we construct four independently fine-tuned models. 
In each evaluation round, one model $\widetilde{M}_j$ is treated as the held-out target, while the remaining three models serve as shadow models. 
The attack classifier is trained only on the labeled member and non-member examples from these three shadow models and is then applied directly to the held-out target, without using any target membership labels or target-specific refitting.
We repeat this procedure four times so that each independently fine-tuned model serves once as the target.

\paragraph{Output-only attack.}
For each shadow model, we extract the output-side representation $\phi_{\mathrm{out}}(x)$ for its member and non-member examples.
The representations from the three shadow models are pooled to train the output-only membership classifier. 
For a candidate $x$ queried against the held-out target, the output-only membership score is
\begin{equation}
    s_{\mathrm{out}}(x)
    =
    g_{\mathrm{out}}\!\left(\phi_{\mathrm{out}}(x)\right).
\end{equation}

\paragraph{Router-augmented attack.}
The router-augmented attack uses the identical shadow-model training and target evaluation protocol, but augments the output-side representation with router telemetry:
\begin{equation}
    \phi_{\mathrm{aug}}(x)
    =
    [\phi_{\mathrm{out}}(x);\phi_R(x)].
\end{equation}
After feature standardization, the membership classifier is trained on the pooled shadow-model examples and applied directly to the held-out target:
\begin{equation}
    s_{\mathrm{aug}}(x)
    =
    g_{\mathrm{aug}}\!\left(\phi_{\mathrm{aug}}(x)\right).
\end{equation}

Because the output-only and router-augmented attacks use the same shadow models, membership labels, classifier family, and held-out target, their performance difference isolates the additional membership signal provided by router telemetry.

\section{Results Across MIA Baselines}
\label{app:baseline_results}

The main experiments use the output-signal ensemble $A_{\mathrm{out}}$ as the primary output-side baseline. 
To examine whether the membership signal provided by router telemetry depends on a particular MIA baseline, we augment each individual output-side baseline with the continuous expert-load representation $\phi_R$ and evaluate it over the full $3\times3$ model--domain grid.

Table~\ref{tab:all_baseline_results} reports the improvements in AUROC and $\operatorname{TPR}@1\%\operatorname{FPR}$ from adding router telemetry. 
Each entry is the difference between the router-augmented attack and the corresponding output-side baseline.

Across the 108 baseline--model--domain combinations, router telemetry improves AUROC in 107 settings and $\operatorname{TPR}@1\%\operatorname{FPR}$ in 104 settings.
Positive gains are observed across likelihood-based, confidence-based, and reference-based MIA baselines. The few cases with negligible or negative gains are concentrated in logit margin and ReCaLL.
Baselines with strong output-only performance, such as LTsig and $A_{\mathrm{out}}$, generally exhibit smaller gains, leaving less headroom for additional improvement.

Overall, router telemetry provides membership information that complements a broad range of output-side MIA baselines. 
This pattern holds for both AUROC and $\operatorname{TPR}@1\%\operatorname{FPR}$, indicating that the additional membership signal is not specific to the output-signal ensemble or to a particular evaluation metric.

\begin{table*}[p]
\centering
\caption{
Improvements from augmenting individual MIA baselines with continuous expert-load telemetry $\phi_R$ across all model--domain settings. 
Each entry reports the difference between the router-augmented attack and its output-only counterpart.
}
\label{tab:all_baseline_results}
\scriptsize
\setlength{\tabcolsep}{4pt}
\renewcommand{\arraystretch}{0.92}
\resizebox{\textwidth}{!}{
\begin{tabular}{lcccccc}
\toprule
& \multicolumn{2}{c}{arXiv}
& \multicolumn{2}{c}{GitHub}
& \multicolumn{2}{c}{Wikipedia} \\
\cmidrule(lr){2-3}
\cmidrule(lr){4-5}
\cmidrule(lr){6-7}
MIA Baseline
& $\Delta$AUROC
& $\Delta$TPR@1\%FPR
& $\Delta$AUROC
& $\Delta$TPR@1\%FPR
& $\Delta$AUROC
& $\Delta$TPR@1\%FPR \\
\midrule

\multicolumn{7}{l}{\textbf{OLMoE-1B-7B}} \\
NLL
& $+0.1550$ & $+0.1187$
& $+0.1428$ & $+0.0557$
& $+0.1695$ & $+0.1417$ \\

Min-K\%
& $+0.1379$ & $+0.0895$
& $+0.1194$ & $+0.0503$
& $+0.1266$ & $+0.1290$ \\

zlib
& $+0.1409$ & $+0.0990$
& $+0.1269$ & $+0.0365$
& $+0.1471$ & $+0.1058$ \\

Min-K\%++
& $+0.1121$ & $+0.0417$
& $+0.0616$ & $+0.0155$
& $+0.1033$ & $+0.0378$ \\

Logit margin
& $+0.0218$ & $+0.0056$
& $+0.0117$ & $+0.0000$
& $+0.0381$ & $+0.0072$ \\

Log-prob. percentiles
& $+0.1350$ & $+0.0844$
& $+0.0927$ & $+0.0345$
& $+0.0946$ & $+0.0845$ \\

Reference loss drop
& $+0.0817$ & $+0.6173$
& $+0.0484$ & $+0.6142$
& $+0.1114$ & $+0.8380$ \\

WBC
& $+0.0205$ & $+0.1916$
& $+0.0590$ & $+0.2837$
& $+0.0101$ & $+0.2583$ \\

EZ
& $+0.0172$ & $+0.1529$
& $+0.0141$ & $+0.2132$
& $+0.0138$ & $+0.4370$ \\

ReCaLL
& $+0.0170$ & $+0.0021$
& $+0.0362$ & $+0.0008$
& $-0.0070$ & $+0.0030$ \\

LTsig
& $+0.0071$ & $+0.0914$
& $+0.0035$ & $+0.0588$
& $+0.0010$ & $+0.0243$ \\

$A_{\mathrm{out}}$
& $+0.0061$ & $+0.0841$
& $+0.0025$ & $+0.0630$
& $+0.0009$ & $+0.0267$ \\

\midrule
\multicolumn{7}{l}{\textbf{Qwen1.5-MoE}} \\
NLL
& $+0.1313$ & $+0.1173$
& $+0.1622$ & $+0.0603$
& $+0.1965$ & $+0.1570$ \\

Min-K\%
& $+0.1191$ & $+0.1006$
& $+0.1349$ & $+0.0653$
& $+0.1516$ & $+0.1432$ \\

zlib
& $+0.1322$ & $+0.1102$
& $+0.1373$ & $+0.0637$
& $+0.1750$ & $+0.1365$ \\

Min-K\%++
& $+0.1104$ & $+0.0441$
& $+0.0953$ & $+0.0272$
& $+0.1085$ & $+0.0470$ \\

Logit margin
& $+0.0264$ & $+0.0037$
& $+0.0240$ & $+0.0043$
& $+0.0539$ & $+0.0085$ \\

Log-prob. percentiles
& $+0.1100$ & $+0.1042$
& $+0.0995$ & $+0.0467$
& $+0.0973$ & $+0.1167$ \\

Reference loss drop
& $+0.0938$ & $+0.6221$
& $+0.0698$ & $+0.4777$
& $+0.0818$ & $+0.7230$ \\

WBC
& $+0.0383$ & $+0.3083$
& $+0.0633$ & $+0.3358$
& $+0.0330$ & $+0.6102$ \\

EZ
& $+0.0200$ & $+0.1922$
& $+0.0222$ & $+0.3205$
& $+0.0195$ & $+0.5418$ \\

ReCaLL
& $+0.0027$ & $-0.0012$
& $+0.0378$ & $-0.0002$
& $+0.0160$ & $-0.0030$ \\

LTsig
& $+0.0072$ & $+0.0986$
& $+0.0038$ & $+0.0912$
& $+0.0026$ & $+0.0530$ \\

$A_{\mathrm{out}}$
& $+0.0056$ & $+0.0733$
& $+0.0033$ & $+0.0757$
& $+0.0022$ & $+0.0407$ \\

\midrule
\multicolumn{7}{l}{\textbf{LLaMA-MoE-v2}} \\
NLL
& $+0.0953$ & $+0.1282$
& $+0.1515$ & $+0.1203$
& $+0.1345$ & $+0.2428$ \\

Min-K\%
& $+0.0827$ & $+0.1043$
& $+0.1001$ & $+0.1148$
& $+0.0931$ & $+0.2175$ \\

zlib
& $+0.0889$ & $+0.1180$
& $+0.1375$ & $+0.0972$
& $+0.1181$ & $+0.2307$ \\

Min-K\%++
& $+0.0826$ & $+0.0568$
& $+0.0934$ & $+0.0743$
& $+0.0866$ & $+0.0910$ \\

Logit margin
& $+0.0406$ & $+0.0099$
& $+0.0391$ & $+0.0113$
& $+0.0632$ & $+0.0118$ \\

Log-prob. percentiles
& $+0.0837$ & $+0.1353$
& $+0.1020$ & $+0.1137$
& $+0.0832$ & $+0.1633$ \\

Reference loss drop
& $+0.0791$ & $+0.4073$
& $+0.0776$ & $+0.2310$
& $+0.1550$ & $+0.7138$ \\

WBC
& $+0.0215$ & $+0.2043$
& $+0.0387$ & $+0.2822$
& $+0.0520$ & $+0.6953$ \\

EZ
& $+0.0108$ & $+0.1224$
& $+0.0177$ & $+0.3388$
& $+0.0282$ & $+0.6515$ \\

ReCaLL
& $+0.0137$ & $+0.0184$
& $+0.0449$ & $+0.0068$
& $+0.0458$ & $+0.0140$ \\

LTsig
& $+0.0033$ & $+0.0479$
& $+0.0057$ & $+0.0740$
& $+0.0039$ & $+0.0675$ \\

$A_{\mathrm{out}}$
& $+0.0026$ & $+0.0445$
& $+0.0054$ & $+0.0937$
& $+0.0030$ & $+0.0648$ \\

\bottomrule
\end{tabular}
}
\end{table*}

\section{Fine-Tuning Regimes}
\label{app:adaptation}

We provide the implementation details for the fine-tuning regimes evaluated in Section~\ref{sec:adaptation}. 
All experiments use OLMoE-1B-7B on arXiv with the same four disjoint membership instances and are evaluated after one epoch of fine-tuning.

\paragraph{Full fine-tuning.}
We update all model parameters, including the router gates, using Adafactor with a learning rate of $8\times10^{-6}$, a batch size of 16, and a maximum sequence length of 256. 
This configuration is identical to the main fine-tuning setup described in Appendix~\ref{app:finetuning}.

\paragraph{Frozen-router fine-tuning.}
We use the same configuration as full fine-tuning but keep the router gate parameters fixed throughout training. 
All other model parameters are updated normally. 
This setting separates the membership signal exposed through router telemetry from changes to the router parameters themselves.

\paragraph{LoRA.}
We apply LoRA~\citep{hu2021loralowrankadaptationlarge} to the attention and expert-MLP projection layers with rank $r=16$ and scaling factor $\alpha=32$. 
We use a learning rate of $2\times10^{-4}$, with approximately 2.2\% of the model parameters trainable. The router gates are not included in the LoRA target modules and therefore remain fixed during fine-tuning.

\paragraph{Instruction tuning.}
We perform full-parameter fine-tuning using an instruction--response format. 
Each arXiv document is converted into the following training example:

\begin{center}
\fbox{
\parbox{0.88\linewidth}{
\ttfamily
\#\#\# Instruction: Write an academic abstract.\\[3pt]
\#\#\# Response: <document>
}
}
\end{center}

The training loss is applied only to the response tokens, while the
instruction tokens are masked from the loss. During feature extraction,
we use the same instruction--response format so that the output-side
signals and router telemetry are computed under the same input context
used during fine-tuning.

\section{Limited Router Telemetry and Shadow-Model Access}
\label{app:access}

We provide the implementation details for the restricted router-telemetry and shadow-model settings evaluated in Section~\ref{sec:access_requirements}.

\subsection{Top-K Router Telemetry}
\label{app:topk}

To restrict the amount of exposed router telemetry, we construct a top-$K$ variant of $\phi_R$ on OLMoE-1B-7B with arXiv.
For each candidate, experts are ranked according to their aggregated document-level routing load. 
We retain the dimensions corresponding to the top-$K$ expert loads and mask the remaining dimensions. 
We evaluate
\begin{equation}
    \text{top-}K \in \{1,2,4,8,16,32\}.
\end{equation}
The restricted representation is then used with the same attack training and evaluation protocol as the full expert-load representation.

This setting is distinct from $\phi_R^{\mathrm{count}}$.
The top-$K$ representation restricts the amount of continuous document-level load information available to the attacker, whereas $\phi_R^{\mathrm{count}}$ is constructed directly from discrete expert selections.

\subsection{Number of Shadow Models}
\label{app:shadow_count}

The main attack uses three shadow models for each held-out target.
To evaluate the effect of reducing shadow-model access, we repeat the OLMoE-1B-7B on arXiv experiment with
$K_{\mathrm{shadow}}\in\{1,2,3\}$.

\begin{wrapfigure}{r}{0.42\textwidth}
    \centering
    \vspace{-8pt}
    \includegraphics[width=\linewidth]{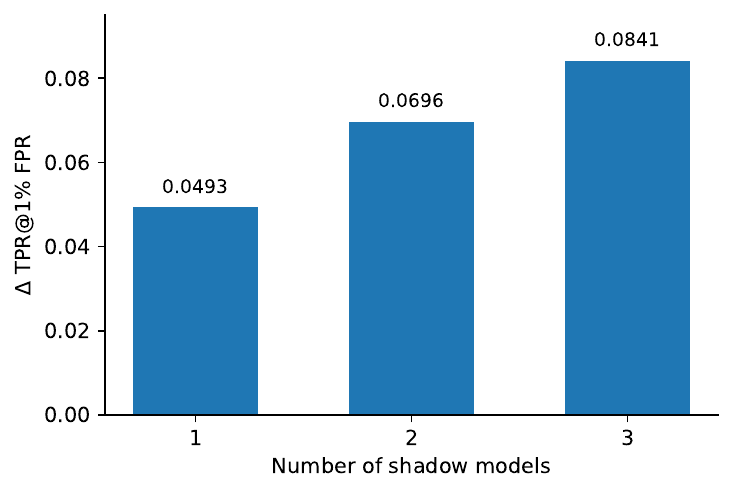}
    \caption{
    Effect of the number of shadow models on OLMoE-1B-7B with arXiv.
    The y-axis reports the improvement in
    $\operatorname{TPR}@1\%\operatorname{FPR}$ over
    $A_{\mathrm{out}}$.
    }
    \label{fig:shadow_count_tpr}
    \vspace{-8pt}
\end{wrapfigure}

For each held-out target and each $K_{\mathrm{shadow}}$, we enumerate all subsets of size $K_{\mathrm{shadow}}$ from the remaining three independently fine-tuned models. 
The membership classifier is trained on the pooled member and non-member examples from each selected subset and evaluated directly on the held-out target. 
Predictions are pooled across all shadow-model subsets and all four held-out targets before computing the evaluation metrics. 
The setting $K_{\mathrm{shadow}}=3$ is identical to the leave-one-out setting used in the main experiments.

As shown in Figure~\ref{fig:shadow_count_tpr}, a single shadow model already yields an improvement of 0.0493 in $\operatorname{TPR}@1\%\operatorname{FPR}$ over $A_{\mathrm{out}}$. 
The gain increases to 0.0696 with two shadow models and 0.0841 with three, showing that the router-augmented attack remains effective even with limited shadow-model access.

\section{Mechanism Analysis}
\label{app:mechanism}

We provide implementation details for the mechanism analysis in the main text. 
The analysis is conducted on OLMoE-1B-7B with arXiv using the same four disjoint membership instances as the main experiments. 
We analyze the pretrained model and the models after one and two epochs of
fine-tuning. 
Hidden states and router weight matrices are recorded at each MoE layer for both member and non-member examples.

\subsection{Router-Subspace Projection}
\label{app:mechanism_projection}

At MoE layer $\ell$, the router maps the hidden state $h_{\ell,t}\in\mathbb{R}^{d}$ of token $t$ to router logits through
\begin{equation}
    z_{\ell,t}
    =
    W_{\ell} h_{\ell,t},
\end{equation}
where $W_{\ell}\in\mathbb{R}^{E\times d}$ is the router weight matrix.
For OLMoE-1B-7B, $d=2048$ and $E=64$. The row space of $W_{\ell}$ therefore defines a low-dimensional subspace of the hidden representation that is directly observed by the router.

To measure the membership information contained in this subspace, we project the hidden states onto the row space of $W_{\ell}$ at each MoE layer and aggregate the projected representations over the input
sequence. 
We then train a membership classifier on the resulting features using five-fold stratified out-of-fold prediction separately for each independently fine-tuned model. 
The reported AUROC is averaged over the four models.

\paragraph{Matched-rank random projection.}
As a control, we construct a random subspace with the same rank as the router subspace at each layer and project the same hidden states onto this subspace. 
The router and random projections therefore use the same layers and the same projection dimensionality. 
This comparison controls for the amount of information retained solely due to dimensionality and tests whether the router subspace captures more membership information than an equally sized generic projection of the hidden state.

\subsection{Frozen-Router Control}
\label{app:mechanism_frozen}

We repeat the complete projection analysis with a frozen-router fine-tuning run. 
The data partitions, training configuration, and evaluation procedure are identical to those of the corresponding full fine-tuning run, except that all router gate parameters are held fixed throughout fine-tuning. The remaining model parameters are updated normally.

For both full fine-tuning and frozen-router fine-tuning, we extract the hidden states and router weight matrices at the pretrained checkpoint and after each fine-tuning epoch. 
We then apply the same router-subspace and matched-rank random-subspace evaluations. 
Because the router weights do not change in the frozen-router setting, this comparison separates membership information exposed through the router subspace from membership information that could arise from direct updates to the router parameters.

\section{Telemetry Perturbation Defenses}
\label{app:defense}

We provide the implementation details for the telemetry perturbation defenses evaluated in the main experiments and report top-1 expert agreement as an additional measure of telemetry utility. 
All experiments use OLMoE-1B-7B on arXiv. 
The perturbations are applied directly to the continuous expert-load representation $\phi_R$ before it is exposed to the attacker, while the model and output-side signals remain unchanged.

\subsection{Perturbation Mechanisms}
\label{app:defense_mechanisms}

\paragraph{Gaussian noise.}
For each telemetry dimension $j$, we add zero-mean Gaussian noise scaled by the empirical standard deviation of that dimension:
\begin{equation}
    \widetilde{r}_{j}
    =
    \max\left(0, r_{j}+\epsilon_{j}\right),
    \qquad
    \epsilon_{j}
    \sim
    \mathcal{N}\left(0,\sigma^{2}s_{j}^{2}\right),
\end{equation}
where $r_j$ is the original expert-load feature and $s_j$ is its empirical standard deviation. 
We evaluate $\sigma\in\{0,0.1,0.25,0.5,1,2,4\}$.

\paragraph{Feature dropout.}
For each candidate input, we independently mask each telemetry dimension with probability $p$:
\begin{equation}
    \widetilde{r}_{j}
    =
    m_j r_j,
    \qquad
    m_j\sim\operatorname{Bernoulli}(1-p).
\end{equation}
We evaluate $p\in\{0,0.25,0.5,0.75,0.9,0.99\}$.
The dropout mask is sampled independently for each candidate input.

\paragraph{Quantization.}
We reduce the precision of the exposed expert loads by quantizing each dimension into a fixed quantization resolution:
\begin{equation}
    \widetilde{r}_{j}
    =
    \frac{1}{B}
    \operatorname{round}\left(B r_j\right),
\end{equation}
where smaller $B$ corresponds to coarser telemetry. 
We evaluate $B\in\{256,32,16,8,4\}$.

For each perturbation level, the perturbed telemetry replaces $\phi_R$ in the same leave-one-out attack pipeline described in Appendix~\ref{app:evaluation}. 
The output-signal ensemble $A_{\mathrm{out}}$ is kept unchanged.

\subsection{Top-1 Expert Agreement}
\label{app:defense_top1}

In addition to the load-correlation metric reported in the main text, we measure whether the perturbed telemetry preserves the identity of the most heavily loaded expert. For each candidate input $i$ and MoE layer $\ell$, top-1 expert agreement is defined as
\begin{equation}
    \operatorname{Agree}_{\mathrm{top1}}
    =
    \frac{1}{NL}
    \sum_{i=1}^{N}
    \sum_{\ell=1}^{L}
    \mathbb{I}
    \left[
        \arg\max_{e} r_{i,\ell,e}
        =
        \arg\max_{e} \widetilde{r}_{i,\ell,e}
    \right],
\end{equation}
where $N$ is the number of candidate inputs and $L$ is the number of MoE layers. 
Higher agreement indicates that the perturbed telemetry better preserves expert-selection information from the original router loads.

\begin{figure*}[t]
    \centering
    \begin{subfigure}[t]{0.32\textwidth}
        \centering
        \includegraphics[width=\linewidth]
        {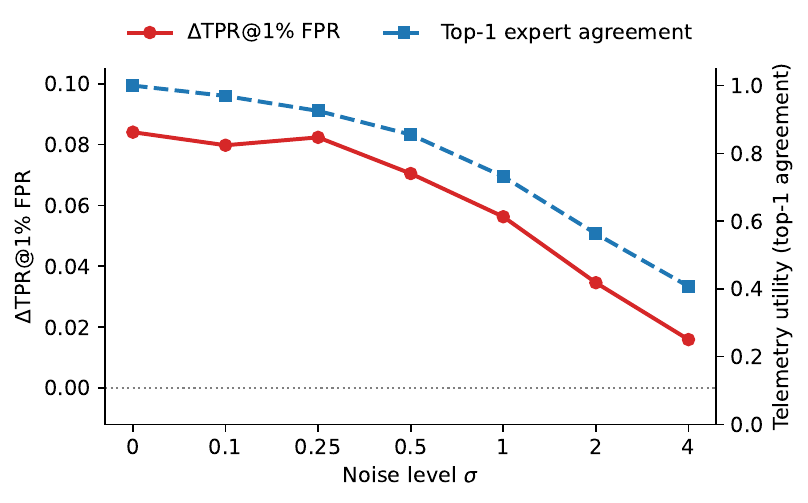}
        \caption{Gaussian noise.}
        \label{fig:defense_gaussian_top1}
    \end{subfigure}
    \hfill
    \begin{subfigure}[t]{0.32\textwidth}
        \centering
        \includegraphics[width=\linewidth]
        {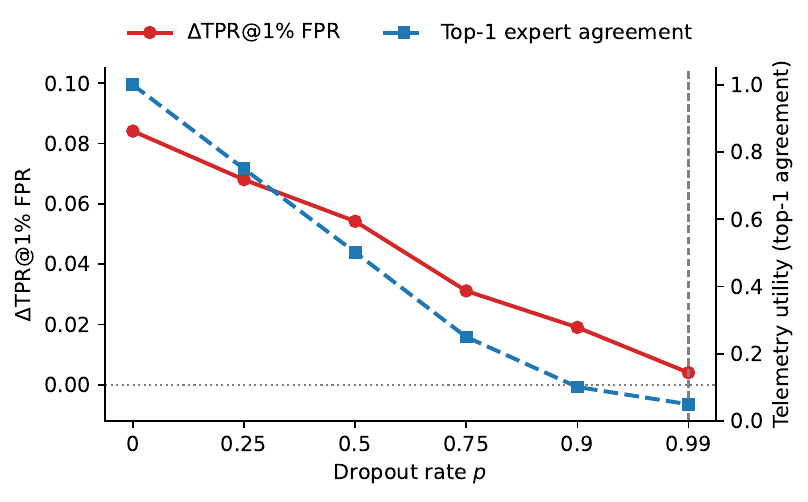}
        \caption{Feature dropout.}
        \label{fig:defense_dropout_top1}
    \end{subfigure}
    \hfill
    \begin{subfigure}[t]{0.32\textwidth}
        \centering
        \includegraphics[width=\linewidth]
        {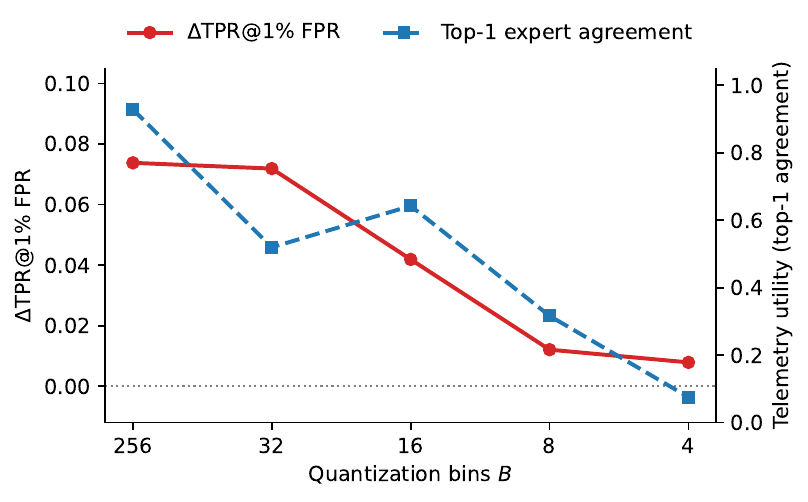}
        \caption{Quantization.}
        \label{fig:defense_quantization_top1}
    \end{subfigure}
    \caption{
    Membership signal and telemetry utility under router-telemetry perturbation on OLMoE-1B-7B with arXiv. 
    The left axis reports the additional $\operatorname{TPR}@1\%\operatorname{FPR}$ from router telemetry, while the right axis reports top-1 expert agreement with the original telemetry.
    }
    \label{fig:defense_top1}
\end{figure*}

Figure~\ref{fig:defense_top1} shows that reducing the membership signal is accompanied by a loss of expert-selection information. 
Under Gaussian noise, both quantities decrease gradually. 
At $\sigma=4$, the additional $\operatorname{TPR}@1\%\operatorname{FPR}$ decreases from 0.0841 to 0.0159, while top-1 expert agreement decreases from 1.000 to 0.407.

Feature dropout produces a sharper reduction in telemetry utility. 
At $p=0.9$, the additional $\operatorname{TPR}@1\%\operatorname{FPR}$ is 0.0190 and top-1 agreement is 0.101. 
Increasing the dropout rate to $p=0.99$ further reduces these values to 0.0040 and 0.050, respectively.

Quantization shows a similar trade-off. With $B=8$, the additional $\operatorname{TPR}@1\%\operatorname{FPR}$ decreases to 0.0121 while top-1 agreement decreases to 0.316. 
At $B=4$, the corresponding values are 0.0079 and 0.076.

Overall, the top-1 agreement results are consistent with the privacy--utility trade-off observed using load correlation in the main text. 
Stronger perturbations reduce the membership signal, but the settings that suppress most of the additional signal also substantially degrade the expert-selection information retained by the exposed telemetry.

\end{document}